\documentclass[10pt,twocolumn,letterpaper]{article}
\usepackage[pagenumbers]{cvpr}

\usepackage{xcolor}         
\definecolor{mydarkred}{rgb}{0.6,0,0}
\definecolor{mydarkgreen}{rgb}{0,0.6,0}
\usepackage{amsfonts}       
\usepackage{nicefrac}       
\usepackage{microtype}      
\usepackage{amsthm}
\usepackage{bm,bbm}

\usepackage{threeparttable}
\usepackage{upgreek}
\usepackage{colortbl}
\usepackage{threeparttable}
\usepackage{multirow}

\usepackage{algorithm}
\usepackage{algorithmic}   

\definecolor{cvprblue}{rgb}{0.21,0.49,0.74}
\usepackage[pagebackref,breaklinks,colorlinks,allcolors=cvprblue]{hyperref}

\def\paperID{*****} 
\def\confName{****}
\def\confYear{2026}

\title{Global-Graph Guided and Local-Graph Weighted Contrastive Learning for Unified Clustering on Incomplete and Noise Multi-View Data}

\author{
Hongqing He$^{1}$,
Jie Xu$^{2,3,*}$,
Wenyuan Yang$^2$,
Yonghua Zhu$^3$,
Guoqiu Wen$^1$,
Xiaofeng Zhu$^{4}$
\\
{$^1$Guangxi Normal University; $^2$University of Electronic Science and Technology of China} \\
{$^3$Singapore University of Technology and Design; $^4$Hainan University}
}

\begin{document}
\maketitle

\let\thefootnote\relax\footnotetext{$^{*}$Corresponding Author.}

\begin{abstract}
Recently, contrastive learning (CL) plays an important role in exploring complementary information for multi-view clustering (MVC) and has attracted increasing attention. Nevertheless, real-world multi-view data suffer from data incompleteness or noise, resulting in rare-paired samples or mis-paired samples which significantly challenges the effectiveness of CL-based MVC. That is, rare-paired issue prevents MVC from extracting sufficient multi-view complementary information, and mis-paired issue causes contrastive learning to optimize the model in the wrong direction. To address these issues, we propose a unified CL-based MVC framework for enhancing clustering effectiveness on incomplete and noise multi-view data. First, to overcome the rare-paired issue, we design a global-graph guided contrastive learning, where all view samples construct a global-view affinity graph to form new sample pairs for fully exploring complementary information. Second, to mitigate the mis-paired issue, we propose a local-graph weighted contrastive learning, which leverages local neighbors to generate pair-wise weights to adaptively strength or weaken the pair-wise contrastive learning. Our method is imputation-free and can be integrated into a unified global-local graph-guided contrastive learning framework. Extensive experiments on both incomplete and noise settings of multi-view data demonstrate that our method achieves superior performance compared with state-of-the-art approaches.
\end{abstract}

\section{Introduction}\label{sec:intro}
Multi-view data refers to the paired data with semantic association and widely exists in practical applications in various forms, such as industry multi-sensor data~\cite{vazquez2020multigraph}, medical multi-omics data~\cite{rappoport2018multi,li2023scbridge}, and intent multi-modal data~\cite{yu2018web}, which can observe the same sample from different views for providing comprehensive understanding for the world.
To utilize multi-view data, multi-view clustering (MVC)~\cite{li2020bipartite,cui2024dual} aims to explore the association information between paired data to produce well-clustered representations, and has been fully investigated over the past two decades~\cite{bickel2004multi,chaudhuri2009multi,liu2018late,xu2022multi}.
Among them, contrastive learning (CL) based MVC has become one of the mainstream deep MVC methods in recent years~\cite{xu2022deep,chen2023deep,xi2025contrastive}, because CL inherently conforms to the learning objectives of MVC.

However, practical multi-view data usually suffer from data incompleteness and data noise, and thus nowadays researchers' focus has shifted to studying incomplete MVC and noise-robust MVC.
To address data incompleteness, many methods usually recover missing data and then perform complete MVC tasks.
For example, COMPLETER~\cite{lin2021completer} combines the idea of CL with data imputation to solve the incomplete MVC problem.
DSIMVC~\cite{tang2022deep} extends CL to the incomplete MVC domain and alleviates the interference caused by missing-view data through a safe multi-view learning mechanism.
DCG~\cite{zhang2025incomplete} employs the diffusion model to generate missing data for incomplete MVC.
To address data noise, existing methods leverage inter-view weighting strategies to balance the optimization across multiple views.
For instance,
\citet{wang2020robust} reduces the impact of unreliable or noise views on learning representations by assigning adaptive weights to each view and projecting the data into a low dimensional subspace.
\citet{xu2024investigating} quantifies the information contribution of each view by learning view specific weights, thereby reducing the interference of noise views and enhancing the impact of valuable views.

\begin{figure*}
  \centering
  \includegraphics[width=1\linewidth]{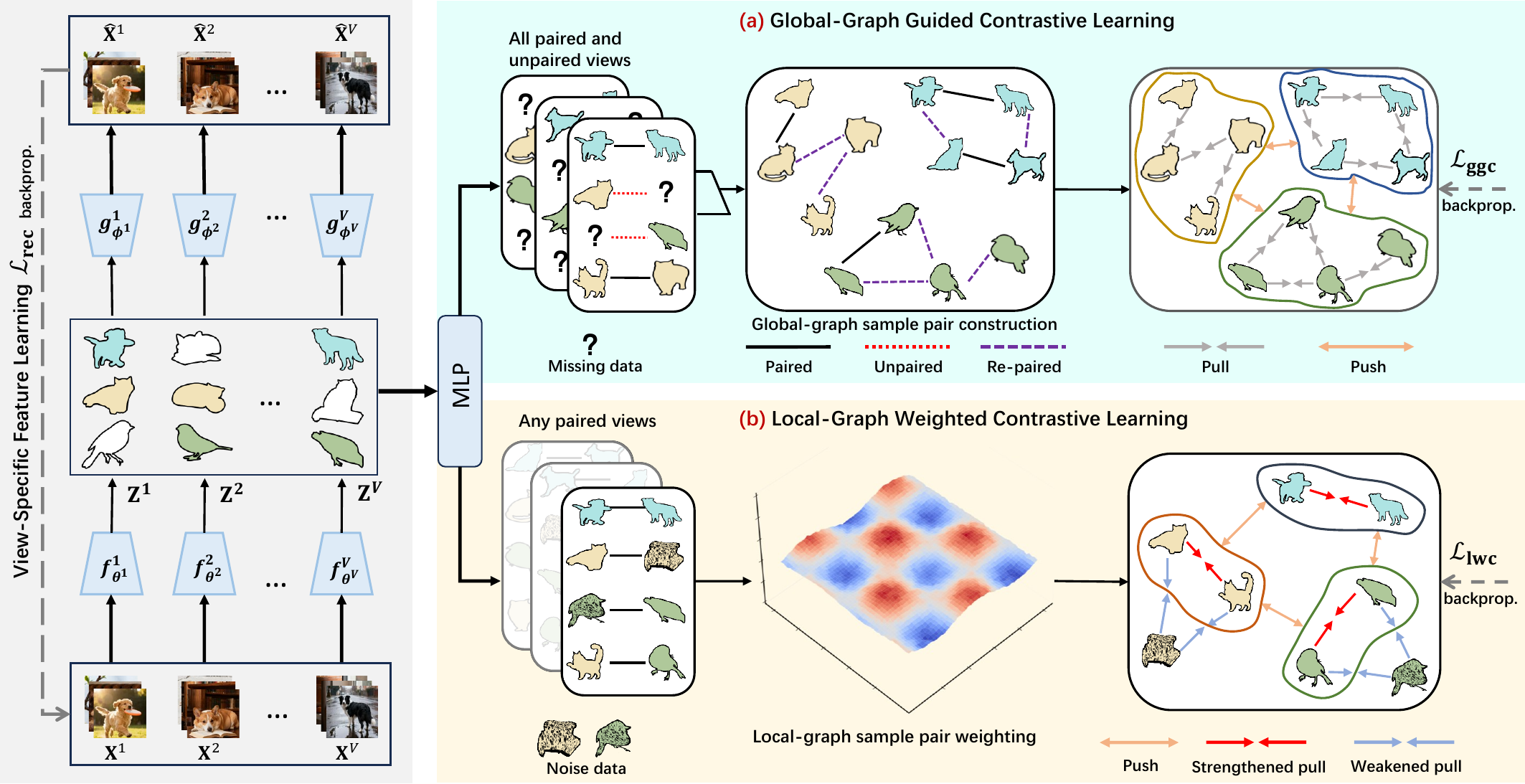}
  \caption{\textbf{Framework Overview of GLC.} Our method consists of two stages: \emph{View-Specific Feature Learning} and \emph{Global-Local Graph-Guided Contrastive Learning}. 
In the first stage, view-specific autoencoders are trained with the reconstruction loss $\mathcal{L}_{\text{rec}}$ to extract view-specific latent features $\{\mathbf{Z}^v\}_{v=1}^{V}$, without imputing the missing data.
In the second stage, we introduce (a) \emph{Global-Graph Guided Contrastive Learning} with loss $\mathcal{L}_{\text{ggc}}$, where all view samples are integrated to build a global-view affinity graph for establishing new sample pairs with semantic association hidden in all views;
and (b) \emph{Local-Graph Weighted Contrastive Learning} with loss $\mathcal{L}_{\text{lwc}}$, which adaptively re-weights cross-view sample pairs based on local feature similarity to suppress the negative effect from noise or unreliable sample pairs.
}
  \label{fig:framework}
\end{figure*}

Despite important advances, previous MVC approaches are still limited by two issues.~1) \emph{Rare-paired issue}:
This arises because most existing CL-based MVC methods~\cite{wang2023self,jin2023deep} are developed under the complete part of multi-view data, neglecting the fact that incomplete part still contain a portion of paired samples whose semantic associations remain under-exploited.
Although several works~\cite{wang2025deep,wu2025imputation} attempt to construct pseudo pairs through data imputation, the imputed data are often unreliable and may introduce additional noise, thereby leading to inaccurate optimization.
Consequently, the rare-paired issue hinders incomplete MVC methods from effectively extracting sufficient complementary information across multiple views.
2) \emph{Mis-paired issue}:
Due to the real-world multi-view data usually contains noise or outliers, the pair between a normal view and the noise view lead to the mis-paired samples. 
These mis-paired samples introduce misleading information and prevent the model from learning true semantic relationships. 
Although some works~\cite{xu2023self,lou2024self} have adopted view-grained weighting strategies to mitigate the impact of noise views, they are limited to adaptively distinguish fine-grained mis-paired samples. 
As a result, the mis-paired issue prompts contrastive learning to optimize the model in the wrong direction, ultimately reducing the quality of learned representations.

To address the aforementioned issues, we propose a Global-Local graph based Contrastive learning (namely GLC as shown in Figure~\ref{fig:framework}) for achieving unified clustering on incomplete and noise multi-view data.
Specifically, we design a \emph{global-graph guided contrastive learning} module, in which samples from all views collaboratively construct a global-view affinity graph to discover indirect semantic correlations and generate additional positive pairs. 
This mechanism enables the model to fully exploit complementary information across all views. 
Then, we introduce a \emph{local-graph weighted contrastive learning} module, which captures local neighborhood relationships and assigns adaptive pair-wise weights based on feature similarity. 
This adaptive weighting contrastive learning effectively strengthens reliable pairs while suppressing noise or unreliable ones, guiding the model toward more robust representation learning.
In this way, this global-local graph-based contrastive learning can provide a unified solution for clustering on incomplete and noise multi-view data.

Compared to previous CL-based MVC methods, the contributions of this work can be summarized as:
\begin{itemize}
    \item We propose a novel global-graph guided contrastive learning module, which can construct potential positive and negative sample pairs across all views, enabling the model to fully exploit the complementary information with semantic association in incomplete multi-view data.
    \item We further design a local-graph weighted contrastive learning module that can assign pair-wise weights based on local neighborhood similarity, which adaptively strengthens reliable positive sample pairs while suppressing noise or unreliable ones.
    \item The propose two modules form a unified contrastive learning framework which is imputation-free and can simultaneously address the rare-paired and mis-paired issues in multi-view learning. Extensive experiments on both incomplete and noise settings of multi-view datasets demonstrate our method's effectiveness.
\end{itemize}

\section{Related Work}
\label{related}
\subsection{Deep Incomplete Multi-View Clustering}
Incomplete multi-view clustering aims to address the problem of missing views that commonly exist in real-world multi-view data. 
Inspired by the expressive power of deep learning, many deep incomplete MVC methods~\cite{wen2020dimc,pu2024adaptive} have been developed. 
The previous work in the literature can be roughly divided into two categories: imputation-based methods and imputation-free methods. 
(i) Imputation-based methods employ various strategies to impute missing view data, and then cluster the completed multi-view dataset.
For example, \citet{wang2021generative} trains Generative Adversarial Networks to recover missing views and then learns the common representations for all views.
CPSPAN~\cite{jin2023deep} completes the representations of missing samples through the transfer of cross-view structural relationships.
GHICMC~\cite{chao2025global} implicitly completes missing view information in the representation layer through global graph propagation and hierarchical information transmission.
(ii) Imputation-free methods typically fully utilize the existing view representations and avoid the inaccurate data recovery in clustering processes.
For instance, \citet{xu2022deep} proposes imputation-free incomplete MVC which independently learns features of each view and mines views' complementarity in a high-dimensional space.
Some work~\cite{xu2024deep} proposes an imputation-free Product-of-Experts fusion method, from which the view-shared representation and clustering assignment are derived.
Considering the view diversity, previous methods usually leverage weighting strategies to achieve the balance optimization across multiple views~\cite{xu2022deep,liu2023adaptive,lou2024self}.
In this work, we follow the imputation-free idea and propose a novel global-local graph-guided contrastive learning method for achieving incomplete and noise-robust MVC. 

\subsection{Multi-View Contrastive Learning}
Multi-view contrastive learning~\cite{tian2020makes} has emerged as a powerful paradigm for learning representations, by treating different views of the same sample as positive pairs and the views of different samples as negative pairs.
Its remarkable success in multi-view clustering~\cite{trosten2021reconsidering,xu2022multi,yang2023dealmvc} stems from the ability to maximize the mutual information between views, and researchers have also combined it with incomplete MVC tasks.
For example, \citet{lin2021completer} propose to optimize the mutual information loss between two views and perform contrastive prediction tasks for incomplete data.
\citet{tang2022deep} first recover the missing data' representation by identifying its neighbors and then optimize a spectral contrastive loss to align representations across views.
\citet{jin2023deep} employs contrastive learning at both the instance and the prototype levels to train the model for incomplete MVC.
Despite the progress, existing methods have overlooked the rare-paired and mis-paired issues, which hinder the effectiveness of contrastive learning in incomplete multi-view scenarios.
In this work, we argue for addressing these two issues to enable contrastive learning applicable for broader applications.

\section{Method}\label{Method}

\textbf{Notations}. We let $\{\mathbf{X}^v \in \mathbb{R}^{N \times D_v}\}^{V}_{v=1}$ denote a multi-view dataset, where $V$ represents the view number and $N$ denotes the sample number.
People usually leverage an indicator matrix $\mathbf{M} \in \{0,1\}^{N \times V}$ to mark the missing data, where $M_{iv} = 0$ means the $v$-th view of the $i$-th sample is missing or noise, otherwise $M_{iv} = 1$.
For each sample with multiple views, e.g., $\{\mathbf{x}_i^1,\mathbf{x}_i^1,\dots,\mathbf{x}_i^V\}$, we have no idea about the data quality of each view or what is noise.
The number of classes among the dataset are assumed to $K$.

\subsection{Preliminaries and Motivation}
\label{motivation}
For deep incomplete multi-view clustering, autoencoder model~\cite{hinton2006reducing} is widely adopted to learning data representations, by optimizing the following reconstruction loss:
\begin{equation}
\label{eq:rec}
\mathcal{L}_{\text{rec}} = 
\sum_{v=1}^V \mathcal{L}_{\text{rec}}^v
=
\sum_{v=1}^V \sum_{i=1}^{N_v}
\left\|
{\mathbf{x}}_i^v - g_{\phi_v}^{v}\big(f_{\theta_v}^{v}({\mathbf{x}}_i^v)\big)
\right\|_2^2,
\end{equation}
where $N_v$ is the number of available data in the $v$-th view, $f_{\theta_v}^{v}$ and $g_{\phi_v}^{v}$ denote the encoder and decoder, $\theta_v$ and $\phi_v$ are model parameters.
The latent representations are $\mathbf{Z}^v = f_{\theta_v}^{v}(\mathbf{X}^v) \in \mathbb{R}^{N_v \times d_z}$, and the reconstructed data are $\hat{\mathbf{X}}^v = g_{\phi_v}^{v}(\mathbf{Z}^v) \in \mathbb{R}^{N_v \times D_v}$.
Furthermore, a contrastive head with multi-layer perceptron (MLP) is stacked on the latent representation to obtain the contrastive feature $\mathbf{H}^v=\text{MLP}(\mathbf{Z}^v) \in \mathbb{R}^{N_v \times d_h}$ for the $v$-th view,
which are further combined with any other view (e.g., $\mathbf{H}^u$) to form the contrastive loss as follows:
\begin{equation}
\label{eq:p}
\mathcal{L}^{v,u}_{con} =
- \, \sum_{P_{ii} \in \mathcal{P}}
\left[
\log
\frac{e^{P_{ii} / \tau}}
{\sum_{P_{ij} \in \mathcal{N}} e^{P_{ij} / \tau}}
\right],
\end{equation}
where $\mathcal{P}$/$\mathcal{N}$ denotes the set of positive/negative sample pair on $\{\mathbf{H}^v,\mathbf{H}^u\}$.
That is, $\{\mathbf{h}^{v}_i, \mathbf{h}^{u}_i\} \in \mathcal{P}$ and $\{\mathbf{h}^{v}_i, \mathbf{h}^{l}_j\}_{j\neq i}^{l=v,u} \in \mathcal{N}$,
$P_{ii}$ and $P_{ij}$ denote the cosine distance between two representations. $\tau$ is a temperature parameter.
In previous methods, multi-view contrastive learning (MVCL) \cite{xu2022multi,lin2022dual,chen2023deep} is conducted by minimizing the sum of reconstruction loss and contrastive loss:
\begin{equation}
\mathcal{L}_{\text{MVCL}} = 
\sum_{v} \mathcal{L}_{\text{rec}}^v + \sum_{v,u}\mathcal{L}^{v,u}.
\end{equation}
\noindent\textbf{Our motivation.} As shown in the above paradigm, MVCL usually considers only non-missing views as positive pairs, while ignoring incomplete views that are unable to be paired due to missing views.
It will introduce the \emph{rare-paired issue} when the missing rate is heavy.
Moreover, MVCL typically treats all sample pairs equally as shown in Eq.~(\ref{eq:p}), which have overlooked the \emph{mis-paired issue} that real-world noise data form incorrect pairs, thereby easily training the model in the wrong direction.

To address these challenges, we propose a novel global-local graph-guided contrastive learning framework (as shown in Figure~\ref{fig:framework}) for unified clustering on incomplete and noise multi-view data.
Specifically, to overcome the rare-paired issue, we design a global-graph guided contrastive learning (Section~\ref{ggg}), where all view samples construct a global-view affinity graph to form new sample pairs for fully exploring complementary information. Meanwhile, to mitigate the mis-paired issue, we propose a local-graph weighted contrastive learning (Section~\ref{lgw}), which leverages local neighbors to generate pair-wise weights to adaptively strength or weaken the pair-wise contrastive learning.
Their details are introduced as following sections.

\subsection{Global-Graph Guided Contrastive Learning}\label{ggg}
To alleviate the rare-paired issue arising from incomplete multi-view data, we introduce a global-graph guided contrastive learning module as illustrated in Figure~\ref{fig:framework}(a).

\begin{figure}[!t]
    \centering
    \begin{subfigure}[b]{0.47\textwidth}
        \includegraphics[width=\textwidth]{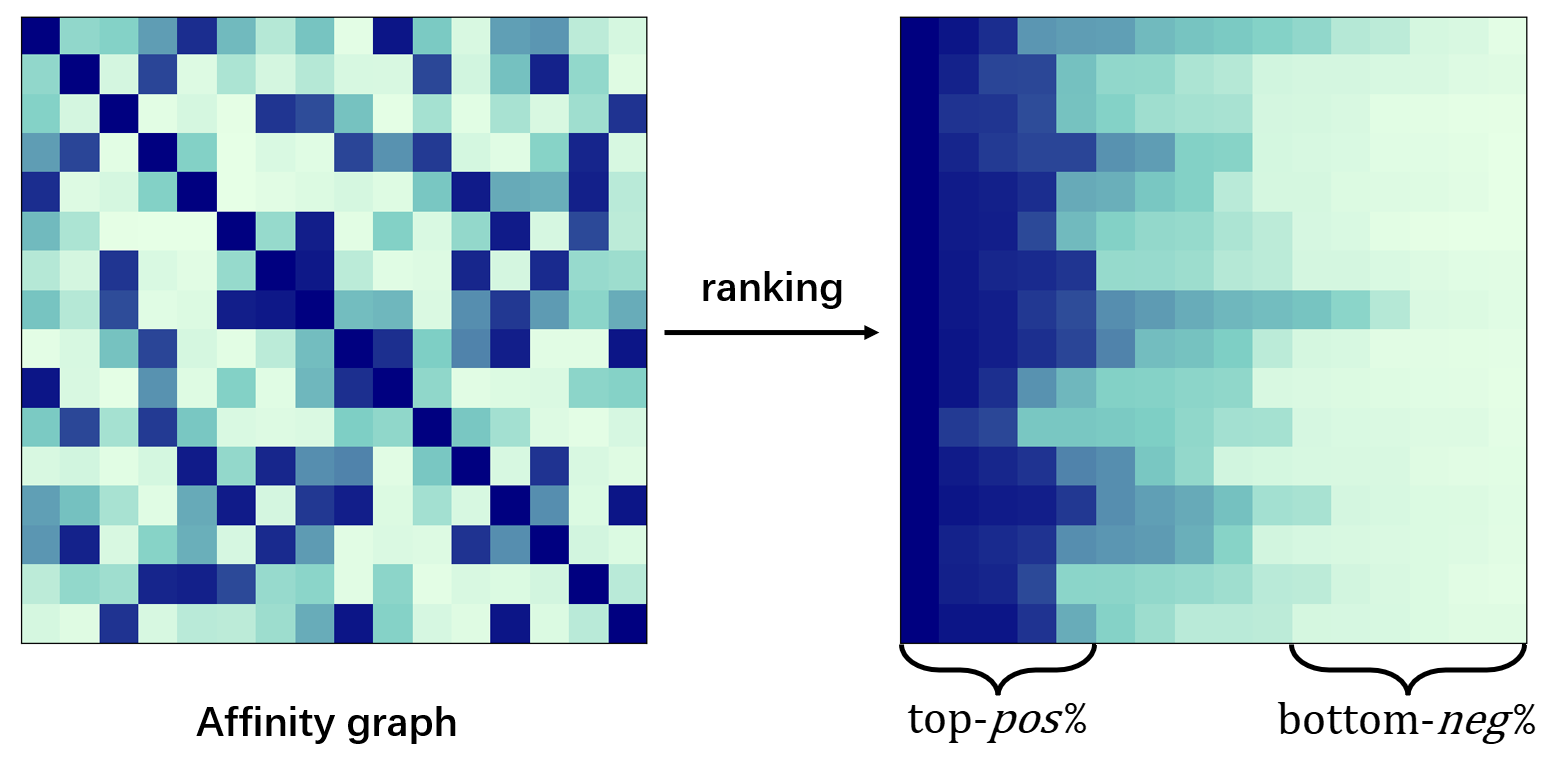}
    \end{subfigure}
    \caption{Illustration of the global-graph sample pair construction. }
    \label{fig:illu}
\end{figure} 

Specifically, we construct a global-view affinity graph $\mathbf{G} \in \mathbb{R}^{N_c \times N_c}$ on the contrastive features of all views.
In $\mathbf{G}$, each edge weight $G_{ij}$ is computed by the cosine similarity between any two features, e.g., $\mathbf{h}_i, \mathbf{h}_j \in \{\mathbf{H}^v\}^{V}_{v=1}$:
\begin{equation}
\label{eq:S}
G_{ij} = \frac{\langle \mathbf{h}_i, \mathbf{h}_j \rangle}{\|\mathbf{h}_i\| \cdot \|\mathbf{h}_j\|} \in \mathbf{G}.
\end{equation}
Based on the affinity graph $\mathbf{G}$, we further adaptively determine positive and negative sample pairs.
For each node feature $\mathbf{h}_i$, the most likely correlated nodes in $\mathbf{G}$ are selected to form positive sample pairs, 
while the weakly correlated nodes are selected to form negative sample pairs:
\begin{equation}
\label{eq:pn}
\begin{cases}
\{\mathbf{h}_i, \mathbf{h}_j\} \in \mathcal{P}_{\text{ggc}}, ~\text{if $G_{ij}>$ top-$pos\%$ values of row $i$},  \\[6pt]
\{\mathbf{h}_i, \mathbf{h}_j\} \in \mathcal{N}_{\text{ggc}}, ~\text{if $G_{ij}<$ bottom-$neg\%$ values of row $i$},
\end{cases}
\end{equation}
where $pos$ and $neg$ are two thresholds to control the proportion of selected samples to all samples.
The sample pair construction is illustrated in Figure~\ref{fig:illu}.
This global graph enables the discovery of potential semantic associations beyond directly paired samples, 
thereby enriching the set of positive pairs to address the rare-paired issue.
Then, our global-graph guided contrastive (GGC) loss is defined as:
\begin{equation}
\label{eq:GGC}
\mathcal{L}^{}_{\text{ggc}} =
- \, \sum_{P_{ii} \in \mathcal{P}_{\text{ggc}}}
\left[
\log
\frac{e^{P_{ii} / \tau}}
{\sum_{P_{ij} \in \mathcal{N}_{\text{ggc}}} e^{P_{ij} / \tau}}
\right],
\end{equation}

Compared with Eq.~(\ref{eq:p}), optimizing our $\mathcal{L}_{\text{ggc}}$ is helpful to discover more complementary information across all views.
Moreover, our method can establish pair associations for the incomplete multi-view data, as thus to alleviate the rare-paired issue in incomplete MVC tasks.

\subsection{Local-Graph Weighted Contrastive Learning}\label{lgw}
To mitigate the mis-paired issue arising from noise multi-view data, we introduce a local-graph weighted contrastive learning module as illustrated in Figure~\ref{fig:framework}(b).

Specifically, we construct a local-view affinity graph $\mathbf{W}^{(u,v)} \in \mathbb{R}^{n \times n}$ on each two features $\{\mathbf{H}^{u}, \mathbf{H}^{v}\}$:
\begin{equation}
\label{eq:W}
W^{(u , v)}_{ij} = \exp\!\Big(-\frac{\|\mathbf{h}^{u}_i - \mathbf{h}^{v}_j\|^2}{\sigma}\Big) \in \mathbf{W}^{(u,v)},
\end{equation}
where $\sigma$ controls the scaling of the distance metric. 
The local-view affinity graph $\mathbf{W}^{(u,v)}$ reflects the geometric affinities between two views.
To further capture indirect semantic associations, we build a high-order local-view graph by propagating similarity through intermediate nodes:
\begin{equation}
\label{eq:G}
\hat{\mathbf{W}}^{(u,v)} = \mathbf{W}^{(u,v)} (\mathbf{W}^{(v,v)})^{T}.
\end{equation}
This high-order graph enriches the structural context of local relationships, enhancing the model’s ability to discover reliable semantic correspondences. 

Based on Eq.~(\ref{eq:p}), we construct the set of positive and negative pairs $\mathcal{P}_{\text{lwc}}, \mathcal{N}_{\text{lwc}}$, and leverage the local-view graph $\hat{\mathbf{W}}^{(u,v)}$ to adaptively re-weight each sample pair according to its semantic correspondences.
For the $u,v$-th views, we formulate local-graph weighted contrastive (LWC) loss as:
\begin{equation}
\label{eq:LWC}
\mathcal{L}^{u,v}_{\text{lwc}} =
- \, \sum_{P_{ii} \in \mathcal{P}_{\text{lwc}}}
\left[
\log
\frac{\hat{W}^{(u , v)}_{ii} e^{P_{ii} / \tau}}
{\sum_{P_{ij} \in \mathcal{N}_{\text{lwc}}} e^{P_{ij} / \tau}}
\right],
\end{equation}
where $\hat{W}^{(u , v)}_{ii} \in \hat{\mathbf{W}}^{(u,v)}$, and the total LWC loss over all views is:
\begin{equation}
\label{eq:LWC_all}
\mathcal{L}_{\text{lwc}}
= \sum_{u=1}^{V} \sum_{v=u+1}^{V} 
\mathcal{L}^{(u,v)}_{\text{lwc}} .
\end{equation}

Minimizing our $\mathcal{L}_{\text{lwc}}$ enables the model to adaptively strengthen reliable positive sample pairs while suppressing noise or unreliable ones, thus improving the model robustness against the mis-paired issue.  

\begin{algorithm}[!t]
\caption{: The training steps of GLC framework}
\label{algorithm}
\small
\begin{algorithmic}[1]  % 加 [1] 显示行号
\STATE \textbf{Input:} Multi-view dataset $\{\mathbf{X}^v \}_{v=1}^V$, indicator matrix $\mathbf{M}$ % number of clusters $C$
\STATE \textbf{Setting:} GGC parameters $pos$ and $neg$, LWC parameter $\sigma$, batch size $|\mathcal{B}|$, learning rate $\eta$, trade-off parameters $\alpha$ and $\beta$
\STATE Pre-train $\{\mathbf{Z}^v\}_{v=1}^V$ by minimizing Eq.~\eqref{eq:rec}
\WHILE{not reaching the maximal iterations}
    \STATE Choose a random mini-batch $\mathcal{B}$ from $\{\mathbf{Z}^v\}_{v=1}^V$
    \STATE Infer $\{\hat{\mathbf{X}}^v,\mathbf{H}^v\}_{v=1}^V$ using autoencoder, contrastive head
    \STATE Compute $\mathbf{G}$, $\mathcal{P}_{\text{ggc}}$ and $\mathcal{N}_{\text{ggc}}$ by Eqs.~\eqref{eq:S} and \eqref{eq:pn}
    \STATE Compute $\hat{\mathbf{W}}^{(u,v)}$, ${\mathcal{P}_{\text{lwc}}}$ and ${\mathcal{N}_{\text{lwc}}}$ by Eqs.~\eqref{eq:G} and \eqref{eq:LWC}
    \STATE Compute the loss function $\mathcal{L}_{\text{GLC}}$ by Eqs.~\eqref{eq:rec}, \eqref{eq:GGC}, and \eqref{eq:LWC}
    \STATE Minimize $\mathcal{L}_{\text{GLC}}$ to update the model with learning rate $\eta$
\ENDWHILE
% \STATE Compute the mean representations by Eq.~\eqref{meanf}
\STATE \textbf{Output:} The mean representations by Eq.~\eqref{meanf}
\end{algorithmic}
\end{algorithm}

\subsection{Loss Function and Clustering}
The final objective function of our proposed Global-Local Graph-guided Contrastive Learning framework (GLC) also integrates the view-specific feature learning, aiming to obtain discriminative and robust representations for clustering. The overall loss function is defined as:
\begin{equation}
\label{eq:total}
\mathcal{L}_{\text{GLC}} = \mathcal{L}_{\text{rec}} + \alpha\mathcal{L}_{\text{ggc}} + \beta\mathcal{L}_{\text{lwc}},
\end{equation}
where $\alpha$ and $\beta$ are trade-off parameters that balance different learning objectives.
To test the clustering performance, we average the learned features of all available samples:
\begin{equation}\label{meanf}
    \hat{\mathbf{h}}_i = \frac{1}{\sum_v{M_{iv}}} \sum_{v} \mathbf{h}_i^{v},~~~~s.t.~M_{iv} = 1.
\end{equation}
Then, we can leverage traditional clustering method such as $K$-means~\cite{mcqueen1967some} to obtain clustering results on $\{\hat{\mathbf{h}}_i\}_{i=1}^N$.

\vspace{+0.05cm}
\noindent\textbf{Complexity analysis.}
The training steps of our GLC framework is shown in Algorithm~\ref{algorithm}.
Let $E$ denote the number of training epochs, $V$ the number of views, $N$ the total number of samples, and $|\mathcal{B}|$ the batch size. 
The view-specific autoencoders perform feature extraction and reconstruction with a cost of $\mathcal{O}(V|\mathcal{B}|)$. 
For Global-Graph Guided Contrastive Learning and Local-Graph Weighted Contrastive Learning, computing pairwise similarities and constructing high-order affinity graphs across all view pairs require $\mathcal{O}(V^2|\mathcal{B}|^2)$ operations per batch. 
The clustering step adds $\mathcal{O}(NC)$, where $C$ is the number of clusters. 
Thus, the total computational complexity for $E$ epochs approximates to $\mathcal{O}(N) + ( EN/|\mathcal{B}|)\mathcal{O}(V^2|\mathcal{B}|^2)$, which scales linearly with the sample size $N$. 

\section{Experiments}
\subsection{Experimental Setup}
\begin{table}[!t]
\caption{Details of multi-view datasets in our experiments.}
\label{dataset}
\vspace{-0.2cm}
\tiny
\centering
\renewcommand\tabcolsep{8.0pt}
\resizebox{0.48\textwidth}{!}{
\begin{tabular}{c|c|c|c}
\toprule[1pt]
% \hline
\textbf{Datasets} & \textbf{\#Samples} & \textbf{\#Views} & \textbf{\#Classes} \\ \hline
DHA & 483 & 2 & 23 \\ \hline
LandUse-21 & 2,100 & 2 & 21 \\ \hline
ProteinFold & 694  & 12 & 27 \\ \hline
ALOI     & 10,800 & 4 & 100 \\ 
\bottomrule[1pt]
\end{tabular}
}
\end{table}
%-----------------------------------------------------------------------------------------%
\begin{table*}[!t]
\caption{\textbf{Clustering Performance Comparison on Incomplete Setting.} We test four datasets with missing rates of $[0.1,0.3,0.5,0.7,1.0]$. Bold and underline denote the best and the second-best results. ``n/a'' signifies that the method could not be executed in that case.}
\label{table:1}
\vspace{-0.2cm}
\centering
\renewcommand\tabcolsep{2.0pt}
\resizebox{\textwidth}{!}{
\begin{tabular}{c|c|cccccccc|cccccccc}
\toprule[2pt]
\multirow{2}{*}{} & \multirow{1}{*}{} & \multicolumn{8}{c|}{ACC} & \multicolumn{8}{c}{NMI} \\
\hline
~~~~~
 &    \rotatebox{45}{rate} 
 &    \rotatebox{45}{DSIMVC~} & \rotatebox{45}{CPSPAN~} & \rotatebox{45}{RPCIC~~} & \rotatebox{45}{SCSL~~~} 
 &    \rotatebox{45}{DCG~~~~} & \rotatebox{45}{GHICMC~} & \rotatebox{45}{FreeCSL} & \rotatebox{45}{\textbf{GLC (ours)}} 
 &    \rotatebox{45}{DSIMVC~} & \rotatebox{45}{CPSPAN~} & \rotatebox{45}{RPCIC~~} & \rotatebox{45}{SCSL~~~} 
 &    \rotatebox{45}{DCG~~~~} & \rotatebox{45}{GHICMC~} & \rotatebox{45}{FreeCSL} & \rotatebox{45}{\textbf{GLC (ours)}} \\
\hline
\multirow{5}{*}{\rotatebox{90}{DHA}} 
& 0.1 & 50.1 & 65.0 & 57.2 & 75.2 & 71.6 & 71.2 & \underline{77.4} & \textbf{84.6} 
      & 64.6 & 77.8 & 68.6 & 82.0 & 80.2 & 76.9 & \underline{85.6} & \textbf{85.9}\\
& 0.3 & 45.0 & 56.9 & 47.8 & 66.2 & 66.8 & 68.2 & \underline{73.3} & \textbf{78.7}
      & 61.3 & 72.7 & 58.4 & 78.6 & 77.2 & 73.6 & \underline{80.5} & \textbf{81.7}\\
& 0.5 & 40.3 & 51.8 & 34.3 & 56.9 & 63.6 & 59.7 & \underline{67.2} & \textbf{75.5}
      & 57.4 & 68.5 & 45.6 & 73.3 & 75.2 & 67.6 & \underline{75.5} & \textbf{78.9}\\
& 0.7 & 40.6 & 44.6 & 23.3 & 46.2 & \underline{58.5} & 51.1 & 55.0 & \textbf{62.4} 
      & 55.0 & 62.4 & 33.7 & 65.2 & \underline{69.9} & 61.8 & 63.9 & \textbf{70.3}\\
& 1.0 & n/a & 32.3 & 17.0 & \underline{38.9} & 23.9 & 35.9 & 32.0 & \textbf{39.4} 
      & n/a & 50.0 & 23.8 & \underline{54.6} & 33.8 & 51.0 & 45.1 & \textbf{56.8}\\
\hline
\multirow{5}{*}{\rotatebox{90}{LandUse-21}} 
& 0.1 & 19.6 & 22.8 & 23.2 & 25.9 & 26.0 & \underline{27.2} & 25.9 & \textbf{27.5} 
      & 20.7 & 30.6 & 29.4 & 28.2 & 29.5 & \textbf{32.0} & 29.3 & \underline{31.0}\\ 
& 0.3 & 18.1 & 21.5 & 18.5 & 24.9 & 25.1 & \textbf{27.6} & 24.8 & \underline{26.7} 
      & 19.7 & 28.9 & 23.0 & 24.9 & 28.6 & \textbf{31.3} & 27.7 & \underline{29.8}\\
& 0.5 & 18.5 & 21.8 & 18.0 & 23.1 & 24.6 & \underline{25.6} & 23.7 & \textbf{26.9} 
      & 20.0 & 28.3 & 22.1 & 23.3 & 26.7 & \underline{29.0} & 26.7 & \textbf{29.8}\\ 
& 0.7 & 17.9 & 21.4 & 16.7 & 22.1 & \underline{24.1} & 23.6 & 23.3 & \textbf{26.5} 
      & 17.7 & 25.8 & 19.4 & 21.3 & 26.3 & \underline{26.6} & 24.3 & \textbf{29.3}\\
& 1.0 & n/a & 16.6 & 17.5 & \underline{17.9} & 12.9 & 17.7 & 16.1 & \textbf{19.1} 
      & n/a & 19.2 & 18.7 & 15.9 & 10.6 & \underline{19.9} & 15.4 & \textbf{21.8}\\
\hline
\multirow{5}{*}{\rotatebox{90}{ProteinFold}} 
& 0.1 & 23.5 & 25.6 & 23.2 & \underline{31.6} & 25.2 & 27.7 & 24.7 & \textbf{32.0} 
      & 32.1 & 32.4 & 32.1 & \underline{38.8} & 33.8 & 36.1 & 34.2 & \textbf{42.7}\\ 
& 0.3 & 22.6 & 24.8 & 20.9 & \underline{28.4} & 23.8 & 25.6 & 25.4 & \textbf{30.2} 
      & 27.9 & 34.1 & 28.0 & \underline{35.9} & 31.3 & 34.8 & 32.3 & \textbf{41.6}\\
& 0.5 & 21.3 & \underline{27.8} & 19.9 & 27.3 & 22.0 & 26.3 & 24.1 & \textbf{30.6} 
      & 25.7 & \underline{36.8} & 29.3 & 34.4 & 28.2 & 34.6 & 30.7 & \textbf{40.8}\\
& 0.7 & 18.2 & 25.8 & 20.4 & \underline{26.6} & 20.8 & 22.9 & 20.8 & \textbf{28.7} 
      & 21.5 & \underline{34.6} & 27.3 & 33.5 & 27.5 & 30.8 & 27.1 & \textbf{38.3}\\
& 1.0 & n/a & \underline{25.5} & 17.4 & 24.5 & 12.9 & 22.1 & 19.7 & \textbf{26.7} 
      & n/a & \underline{32.8} & 21.7 & 29.7 & 13.0 & 28.1 & 25.7 & \textbf{35.2}\\
\hline
\multirow{5}{*}{\rotatebox{90}{ALOI}} 
& 0.1 & 41.6 & 67.3 & 76.3 & 55.9 & 60.5 & 69.1 & \underline{87.1} & \textbf{89.2} 
      & 68.6 & 84.0 & 88.1 & 72.1 & 85.3 & 81.5 & \underline{93.0} & \textbf{93.8}\\
& 0.3 & 39.7 & 67.7 & 68.4 & 42.4 & 58.1 & 63.8 & \underline{84.0} & \textbf{88.5} 
      & 65.6 & 84.5 & 84.6 & 62.0 & 83.9 & 76.6 & \underline{90.8} & \textbf{93.0}\\
& 0.5 & 39.0 & 65.6 & 56.9 & 32.4 & 52.1 & 58.8 & \underline{81.7} & \textbf{87.6} 
      & 64.2 & 83.0 & 80.0 & 54.9 & 80.2 & 71.6 & \underline{88.4} & \textbf{91.9}\\
& 0.7 & 35.2 & 67.7 & 59.8 & 27.6 & 43.9 & 55.1 & \underline{75.5} & \textbf{85.4} 
      & 60.6 & 83.6 & 80.8 & 51.4 & 74.1 & 67.2 & \underline{84.5} & \textbf{90.5}\\
& 1.0 & n/a & \underline{66.7} & 45.5 & 27.2 & 8.9 & 44.7 & 48.1 & \textbf{82.8} 
      & n/a & \underline{83.3} & 69.0 & 50.6 & 23.3 & 57.9 & 67.9 & \textbf{87.7}\\

% \hline
\bottomrule[2pt]
\end{tabular}
}
\end{table*}

\begin{table*}[!t]
\caption{\textbf{Clustering Performance Comparison on Noise Setting.} We test four datasets with the noise rates of $[0.1,0.3,0.5,0.7,1.0]$.}
\label{table:2}
\vspace{-0.2cm}
\centering
\renewcommand\tabcolsep{2.0pt}
\resizebox{\textwidth}{!}{
\begin{tabular}{c|c|cccccccc|cccccccc}
\toprule[2pt]
\multirow{2}{*}{} & \multirow{1}{*}{} & \multicolumn{8}{c|}{ACC} & \multicolumn{8}{c}{NMI} \\
\hline
~~~~~
 &    \rotatebox{45}{rate} 
 &    \rotatebox{45}{DSIMVC~} & \rotatebox{45}{CPSPAN~} & \rotatebox{45}{RPCIC~~} & \rotatebox{45}{SCSL~~~} 
 &    \rotatebox{45}{DCG~~~~} & \rotatebox{45}{GHICMC~} & \rotatebox{45}{FreeCSL} & \rotatebox{45}{\textbf{GLC (ours)}} 
 &    \rotatebox{45}{DSIMVC~} & \rotatebox{45}{CPSPAN~} & \rotatebox{45}{RPCIC~~} & \rotatebox{45}{SCSL~~~} 
 &    \rotatebox{45}{DCG~~~~} & \rotatebox{45}{GHICMC~} & \rotatebox{45}{FreeCSL} & \rotatebox{45}{\textbf{GLC (ours)}} \\
\hline
\multirow{5}{*}{\rotatebox{90}{DHA}} 
& 0.1 & 60.3 & 60.4 & 60.3 & 64.9 & 68.3 & 71.1 & \underline{74.6} & \textbf{83.9}
      & 72.4 & 77.8 & 74.6 & 74.9 & \underline{79.1} & 76.9 & \textbf{85.2} & \textbf{85.2}\\
& 0.3 & 59.4 & 40.9 & 47.7 & \underline{75.5} & 67.3 & 58.5 & 73.8 & \textbf{81.3}
      & 70.9 & 66.1 & 67.2 & 82.5 & 78.6 & 66.6 & \underline{83.0} & \textbf{83.6}\\
& 0.5 & 56.9 & 37.2 & 38.2 & 73.6 & 66.6 & 52.6 & \underline{74.7} & \textbf{80.5}
      & 69.4 & 61.1 & 59.9 & \underline{81.6} & 77.2 & 64.6 & 81.2 & \textbf{82.7}\\
& 0.7 & 53.0 & 36.6 & 40.7 & 72.0 & 65.2 & 46.7 & \underline{73.3} & \textbf{78.0}
      & 66.1 & 53.1 & 59.4 & \underline{80.4} & 76.2 & 59.4 & 80.1 & \textbf{81.4}\\
& 1.0 & 45.5 & 34.3 & 43.4 & \underline{70.2} & 63.4 & 38.3 & 61.3 & \textbf{72.2}
      & 59.7 & 49.8 & 58.6 & \underline{76.2} & 74.5 & 51.6 & 73.0 & \textbf{77.9}\\
\hline
\multirow{5}{*}{\rotatebox{90}{LandUse-21}} 
& 0.1 & 19.8 & 18.4 & 19.6 & 26.1 & 26.2 & 25.9 & \underline{26.8} & \textbf{27.3}
      & 20.7 & 24.2 & 25.8 & 26.9 & 29.5 & \underline{30.6} & 29.7 & \textbf{31.6}\\
& 0.3 & 18.9 & 16.2 & 16.1 & 24.7 & 24.9 & 25.4 & \underline{26.3} & \textbf{27.8}
      & 19.8 & 18.9 & 18.4 & 25.8 & 27.9 & 27.7 & \underline{28.7} & \textbf{31.6}\\
& 0.5 & 19.2 & 14.0 & 13.7 & 22.8 & 24.8 & \underline{25.0} & 22.8 & \textbf{27.4}
      & 20.3 & 12.3 & 12.9 & 21.5 & \underline{27.3} & 26.7 & 24.7 & \textbf{30.9}\\
& 0.7 & 17.6 & 12.4 & 13.1 & 23.1 & \underline{25.1} & 24.8 & 21.0 & \textbf{26.1}
      & 18.1 & 10.4 & 10.6 & 23.5 & \underline{27.7} & 25.6 & 21.7 & \textbf{30.2}\\
& 1.0 & 17.3 & 11.4 & 13.7 & 21.9 & 21.1 & \underline{22.5} & 14.6 & \textbf{24.7}
      & 16.9 & 7.1 & 11.1 & 21.2 & \underline{24.1} & 22.3 & 12.6 & \textbf{27.4}\\
\hline
\multirow{5}{*}{\rotatebox{90}{ProteinFold}} 
& 0.1 & 19.8 & 18.9 & 19.8 & \underline{29.9} & 22.0 & 25.5 & 22.3 & \textbf{31.8}
      & 20.7 & 16.0 & 17.5 & \underline{38.2} & 30.6 & 35.8 & 31.8 & \textbf{43.7}\\
& 0.3 & 18.9 & 13.3 & 14.6 & \underline{27.8} & 21.6 & 25.7 & 21.1 & \textbf{31.7}
      & 19.8 & 8.8 & 10.6 & \underline{34.7} & 27.5 & 33.6 & 27.8 & \textbf{42.9}\\
& 0.5 & 19.2 & 13.0 & 13.0 & \underline{27.4} & 20.7 & 24.7 & 18.5 & \textbf{31.5}
      & 20.3 & 9.2 & 9.5 & 32.8 & 25.7 & \underline{33.1} & 24.6 & \textbf{42.0}\\
& 0.7 & 17.6 & 13.0 & 12.8 & \underline{24.9} & 20.9 & 23.1 & 17.5 & \textbf{30.5}
      & 18.1 & 8.2 & 8.3 & 30.1 & 27.7 & \underline{31.8} & 22.5 & \textbf{40.9}\\
& 1.0 & 17.3 & 12.9 & 12.7 & 20.8 & 19.7 & \underline{22.6} & 13.7 & \textbf{29.5}
      & 16.9 & 10.6 & 10.0 & 26.2 & 27.3 & \underline{30.3} & 18.5 & \textbf{40.3}\\
\hline
\multirow{5}{*}{\rotatebox{90}{ALOI}} 
& 0.1 & 39.7 & 49.1 & 50.2 & 46.4 & 58.2 & 69.1 & \textbf{89.4} & \underline{88.5}
      & 67.3 & 70.7 & 73.8 & 67.5 & 82.4 & 81.5 & \textbf{94.2} & \underline{93.2}\\
& 0.3 & 39.4 & 24.5 & 25.1 & 43.7 & 51.0 & 63.8 & \underline{88.7} & \textbf{90.2}
      & 65.9 & 49.9 & 53.0 & 59.9 & 73.5 & 76.6 & \underline{92.9} & \textbf{93.3}\\
& 0.5 & 38.8 & 13.6 & 14.1 & 35.3 & 40.2 & 58.8 & \underline{66.9} & \textbf{88.6}
      & 63.8 & 35.4 & 37.1 & 52.4 & 64.4 & 71.6 & \underline{75.6} & \textbf{91.5}\\
& 0.7 & 34.8 & 10.1 & 11.1 & 22.1 & 36.6 & \underline{55.1} & 44.6 & \textbf{86.1}
      & 60.5 & 28.2 & 32.5 & 38.3 & 59.8 & \underline{67.2} & 57.2 & \textbf{89.0}\\
& 1.0 & 23.7  & 6.9 & 10.1 & 5.7 & 32.3 & \underline{44.7} & 21.5 & \textbf{79.9}
      & 52.6  & 20.1 & 29.3 & 15.7 & 56.0 & \underline{57.9} & 34.5 & \textbf{83.6}\\
% \hline
\bottomrule[2pt]
\end{tabular}
}
\end{table*}

\begin{table*}[!t]
\caption{\textbf{Clustering Performance Comparison on Incomplete + Noise Setting.} We test four datasets with rates of $[0.1,0.3,0.5,0.7,1.0]$.
}
\label{table:3}
\vspace{-0.2cm}
\centering
\renewcommand\tabcolsep{2.0pt}
\resizebox{\textwidth}{!}{
\begin{tabular}{c|c|cccccccc|cccccccc}
\toprule[2pt]
% \hline
\multirow{2}{*}{} & \multirow{1}{*}{} & \multicolumn{8}{c|}{ACC} & \multicolumn{8}{c}{NMI} \\
\hline
~~~~~
 &    \rotatebox{45}{rate} 
 &    \rotatebox{45}{DSIMVC~} & \rotatebox{45}{CPSPAN~} & \rotatebox{45}{RPCIC~~} & \rotatebox{45}{SCSL~~~} 
 &    \rotatebox{45}{DCG~~~~} & \rotatebox{45}{GHICMC~} & \rotatebox{45}{FreeCSL} & \rotatebox{45}{\textbf{GLC (ours)}} 
 &    \rotatebox{45}{DSIMVC~} & \rotatebox{45}{CPSPAN~} & \rotatebox{45}{RPCIC~~} & \rotatebox{45}{SCSL~~~} 
 &    \rotatebox{45}{DCG~~~~} & \rotatebox{45}{GHICMC~} & \rotatebox{45}{FreeCSL} & \rotatebox{45}{\textbf{GLC (ours)}} \\
\hline
\multirow{5}{*}{\rotatebox{90}{DHA}} 
& 0.1 & 60.0 & 60.0 & 51.4 & 72.3 & 60.9 & 68.6 & \underline{74.3} & \textbf{77.5}
      & 70.6 & 77.1 & 66.1 & 81.4 & 79.5 & 74.1 & \underline{82.4} & \textbf{82.7}\\
& 0.3 & 55.1 & 46.1 & 35.9 & 60.9 & 56.0 & 50.4 & \underline{68.1} & \textbf{76.2}
      & 65.6 & 68.4 & 49.5 & 74.4 & 75.5 & 59.2 & \underline{75.8} & \textbf{79.8}\\
& 0.5 & 48.2 & 45.6 & 26.0 & 47.6 & 50.2 & 42.1 & \underline{57.4} & \textbf{65.2}
      & 61.4 & 64.2 & 37.4 & 63.3 & \underline{67.1} & 50.4 & 64.3 & \textbf{71.4}\\
& 0.7 & 41.3 & 40.4 & 17.2 & 34.4 & \underline{49.4} & 28.8 & 41.3 & \textbf{54.2}
      & 54.4 & \textbf{63.9} & 25.8 & 51.2 & 63.1 & 37.5 & 48.5 & \underline{62.8}\\
& 1.0 & n/a & 28.8 & 18.1 & \underline{31.9} & 18.9 & 17.8 & 26.2 & \textbf{34.3}
      & n/a & 41.1 & 26.2 & \underline{43.0} & 23.7 & 34.2 & 23.9 & \textbf{47.8}\\
\hline
\multirow{5}{*}{\rotatebox{90}{LandUse-21}} 
& 0.1 & 19.3 & 18.8 & 19.9 & \underline{26.2} & 26.1 & 25.8 & 25.5 & \textbf{27.7}
      & 20.7 & 24.5 & 24.3 & 27.6 & 28.8 & \underline{30.0} & 28.7 & \textbf{31.1}\\
& 0.3 & 18.2 & 14.3 & 13.3 & 24.7 & \underline{25.2} & 24.5 & 24.3 & \textbf{26.8}
      & 19.0 & 14.7 & 12.0 & 24.3 & \underline{27.8} & 25.5 & 25.0 & \textbf{29.8}\\
& 0.5 & 18.1 & 12.0 & 11.6 & 23.8 & \underline{24.0} & 22.5 & 21.3 & \textbf{25.9}
      & 18.9 & 9.7 & 8.0 & 22.0 & \underline{25.1} & 22.0 & 19.0 & \textbf{28.6}\\
& 0.7 & 16.2 & 11.4 & 11.2 & 20.1 & \underline{21.9} & 18.6 & 16.7 & \textbf{24.3}
      & 15.4 & 7.3 & 7.2 & 18.0 & \underline{22.2} & 16.2 & 14.2 & \textbf{25.7}\\
& 1.0 & n/a & 11.1 & 11.7 & \textbf{17.1} & 11.9 & \underline{15.3} & 12.1 & \textbf{17.1}
      & n/a & 5.5 & 7.3 & \underline{15.9} & 9.0 & 11.6 & 7.8 & \textbf{17.3}\\
\hline
\multirow{5}{*}{\rotatebox{90}{ProteinFold}} 
& 0.1 & 22.5 & 16.6 & 16.5 & \underline{30.4} & 21.6 & 21.8 & 24.2 & \textbf{31.3}
      & 30.5 & 12.3 & 13.8 & \underline{38.5} & 29.4 & 24.8 & 33.4 & \textbf{42.2}\\
& 0.3 & 21.8 & 13.6 & 13.5 & \underline{28.2} & 19.7 & 21.5 & 20.4 & \textbf{29.7}
      & 27.2 & 8.6 & 9.3 & \underline{36.5} & 25.1 & 23.3 & 24.8 & \textbf{39.7}\\
& 0.5 & 18.9 & 13.0 & 12.8 & \underline{26.8} & 18.7 & 20.8 & 17.2 & \textbf{29.3}
      & 24.2 & 8.0 & 7.9 & \underline{33.0} & 24.0 & 20.8 & 21.2 & \textbf{38.4}\\
& 0.7 & 16.7 & 13.0 & 12.6 & \textbf{26.7} & 18.0 & 20.5 & 15.5 & \underline{26.1}
      & 22.7 & 8.6 & 7.8 & \underline{33.1} & 22.9 & 20.1 & 18.9 & \textbf{35.7}\\
& 1.0 & n/a & 12.4 & 12.6 & \underline{24.1} & 12.3 & 18.2 & 13.5 & \textbf{24.2}
      & n/a & 10.1 & 10.1 & \underline{30.2} & 12.5 & 18.4 & 16.2 & \textbf{31.8}\\
\hline
\multirow{5}{*}{\rotatebox{90}{ALOI}} 
& 0.1 & 42.2 & 48.5 & 51.0 & 49.9 & 46.7 & 65.4 & \textbf{88.1} & \underline{87.4}
      & 69.2 & 69.8 & 72.8 & 67.2 & 74.9 & 79.2 & \textbf{93.0} & \underline{92.4}\\
& 0.3 & 36.6 & 27.2 & 27.7 & 32.5 & 39.6 & 56.2 & \underline{72.2} & \textbf{86.1}
      & 63.8 & 51.1 & 53.9 & 51.3 & 64.6 & 71.0 & \underline{80.0} & \textbf{89.3}\\
& 0.5 & 29.1 & 14.1 & 15.0 & 16.8 & 31.3 & \underline{48.1} & 39.5 & \textbf{79.4}
      & 57.8 & 35.2 & 38.6 & 37.1 & 54.7 & \underline{61.1} & 53.2 & \textbf{82.9}\\
& 0.7 & 22.8 & 10.1 & 11.5 & 13.1 & 22.8 & \underline{39.0} & 21.3 & \textbf{69.8}
      & 49.1 & 26.8 & 32.9 & 31.6 & 43.5 & \underline{50.6} & 37.0 & \textbf{74.7}\\
& 1.0 & n/a & 7.3 & 12.2 & 7.8 & 3.6 & \underline{26.2} & 12.1 & \textbf{48.5}
      & n/a & 18.8 & 32.4 & 21.4 & 8.6 & \underline{36.7} & 27.7 & \textbf{55.1}\\
% \hline
\bottomrule[2pt]
\end{tabular}
}
\end{table*}

\begin{table}[!t]
\caption{\textbf{Ablation Studies on Loss Components} across four datasets. I: incomplete, N: noise, I+N: incomplete and noise.}
\label{table:4}
\vspace{-0.2cm}
\centering
\renewcommand\tabcolsep{2.5pt}
\resizebox{0.48\textwidth}{!}{
\begin{tabular}{c|ccc|cc|cc|cc|cc}
\toprule[2pt]
\multicolumn{1}{c|}{}
&\multicolumn{3}{c|}{Loss} 
& \multicolumn{2}{c|}{DHA} 
& \multicolumn{2}{c|}{Landuse-21} 
& \multicolumn{2}{c|}{ProteinFold} 
& \multicolumn{2}{c}{ALOI} \\
\hline
& $\mathcal{L}_{\text{rec}}$ & $\mathcal{L}_{\text{ggc}}$ & $\mathcal{L}_{\text{lwc}}$
& ACC & NMI  
& ACC & NMI 
& ACC & NMI 
& ACC & NMI \\
\hline
\multirow{3}{*}{\rotatebox{90}{I}} 
& $\checkmark$ &  &  
& 31.7 & 47.2 
& 15.1 & 16.0 
& 17.0 & 18.3  
& 29.7 & 49.5 \\

& $\checkmark$ & $\checkmark$ &  
& 35.9 & 52.9 
& 23.3 & 24.7 
& 17.1 & 20.6  
& 32.9 & 54.1 \\

& $\checkmark$ & $\checkmark$ & $\checkmark$ 
& \textbf{75.5} & \textbf{78.9} 
& \textbf{26.9} & \textbf{29.8}  
& \textbf{30.6} & \textbf{40.8}  
& \textbf{87.6} & \textbf{92.0} \\

\hline
\multirow{3}{*}{\rotatebox{90}{N}}
& $\checkmark$ &  &  
& 44.5 & 54.9  
& 22.0 & 24.9 
& 17.4 & 25.2 
& 27.5 & 42.8 \\

& $\checkmark$ & $\checkmark$ &  
& 46.8 & 58.8 
& 25.6 & 27.3
& 19.7 & 28.2  
& 33.4 & 48.2 \\

& $\checkmark$ & $\checkmark$ & $\checkmark$ 
& \textbf{80.5} & \textbf{82.7}  
& \textbf{27.4} & \textbf{30.9}  
& \textbf{31.5} & \textbf{42.0}  
& \textbf{88.6} & \textbf{91.5} \\

\hline
\multirow{3}{*}{\rotatebox{90}{I+N}}
& $\checkmark$ &  &  
& 31.1 & 44.1 
& 15.4 & 15.5 
& 14.6 & 15.7  
& 13.9 & 29.7 \\

& $\checkmark$ & $\checkmark$ &  
& 34.9 & 49.1  
& 20.8 & 21.7 
& 16.5 & 19.0 
& 17.6 & 36.1 \\

& $\checkmark$ & $\checkmark$ & $\checkmark$ 
& \textbf{65.2} & \textbf{71.4}  
& \textbf{25.9} & \textbf{28.6}  
& \textbf{29.3} & \textbf{38.4}  
& \textbf{79.4} & \textbf{82.9} \\

\bottomrule[2pt]
\end{tabular}
}
\end{table}

\noindent
\textbf{Datasets.}
Our experiments employ four open-source multi-view datasets. 
Their information is shown in Table~\ref{dataset}, where DHA~\cite{lin2012human} is a depth-included human action dataset where each action has RGB and depth features; 
LandUse-21~\cite{yang2010bag} consists of aerial images cropped from various regions across the United States, where each sample is represented by two views, i.e., PHOG and LBP features; ProteinFold~\cite{damoulas2008probabilistic} is a bioinformatics dataset in which each sample is represented by 12 different feature views, including physicochemical properties, sequence information, secondary structure, volume, polarity, and substitution matrix features.
ALOI~\cite{du2021deep} is an image dataset that extracts HSB, RGB, Colorsim, and Haralick features from images to construct multi-view data.

To ensure a fair and consistent evaluation across these datasets, we adopt the same experimental protocols as in \cite{xu2022deep,tang2022deep}.
Specifically, we construct three experimental settings.
In the \emph{Incomplete Setting}, incomplete multi-view samples are generated by randomly removing views while ensuring that each sample retains at least one available view.
In the \emph{Noise Setting}, Gaussian noise with mean 0 and standard deviation 0.4 is randomly added to each view, following the same random perturbation strategy used in the incomplete setting.
For the combined \emph{Incomplete and Noise Setting}, we first randomly inject Gaussian noise as in the noise setting, and then apply view-missing perturbations in the same manner as in the incomplete setting.

\vspace{+0.05cm}
\noindent
\textbf{Comparison methods.}
The comparison methods include DSIMVC~\cite{tang2022deep}, CPSPAN~\cite{jin2023deep}, RPCIC~\cite{yuan2024robust}, SCGL~\cite{liu2024sample}, DCG~\cite{zhang2025incomplete}, GHICMC~\cite{chao2025global} and FreeCSL~\cite{dai2025imputation}. 
We leverage two metrics for evaluation, i.e., clustering accuracy (ACC), normalized mutual information (NMI), and report the mean results with standard deviation of 5 runs.

\vspace{+0.05cm}
\noindent
\textbf{Implementation details.} 
All experiments are conducted on a single GeForce RTX 2080 Ti GPUs (12 GB cache) using PyTorch~\cite{paszke2019pytorch} v1.12.0 with CUDA 10.2.
For our GLC, the encoder network is configured as $\mathbf{X}^v \rightarrow 500 \rightarrow 500 \rightarrow 2000 \rightarrow \mathbf{Z}^v$, 
the decoder network as $\mathbf{Z}^v \rightarrow 500 \rightarrow 500 \rightarrow 2000 \rightarrow \hat{\mathbf{X}}^v$, 
the feature MLP as $\mathbf{Z}^v \rightarrow \mathbf{H}^v$ with ReLU adopted as the activation function. 
For all views, the dimensions of $\mathbf{Z}^v$ and $\mathbf{H}^v$ are set to 512 and 128, respectively.
Across all datasets, we consistently use a batch size of 256, and set the temperature coefficients $\tau$ to 0.5 in contrastive learning. 
The model is optimized by Adam~\cite{kingma2014adam}.  

\subsection{Comparison Experiments.}
Tables~\ref{table:1}, \ref{table:2}, \ref{table:3} respectively show the effectiveness of our GLC and comparison methods on clustering in incomplete, noise, and incomplete + noise settings.

\vspace{+0.05cm}
\noindent
\textbf{Clustering performance on incomplete setting.}
We first evaluate the proposed GLC framework on incomplete multi-view clustering tasks, and the results are reported in Table~\ref{table:1}.
It can be observed that GLC consistently outperforms all comparison methods under different missing rates.
For instance, on the ALOI dataset, GLC achieves an average improvement of 11.7\% in ACC compared to the recent contrastive-based incomplete MVC method FreeCSL.
Notably, when the missing rate reaches 1.0, GLC still surpasses FreeCSL by 34.7\%.
This demonstrates that the proposed GGC alleviates rare pairing problems by constructing global semantic associations between views, ensuring the effectiveness of GLC in severely incomplete situations.

\vspace{+0.05cm}
\noindent
\textbf{Clustering performance on noise setting.}
To further examine the influence of noise, we conduct experiments on noise multi-view datasets, as summarized in Table~\ref{table:2}.
When noise corruption exists, most previous incomplete MVC methods exhibit noticeable performance degradation.
In contrast, our proposed GLC demonstrates remarkable robustness and achieves the best performance across all datasets.
For example, on the ProteinFold dataset, as the noise rate increases from 0.1 to 1.0, the ACC of the second-best method drops by 9.1\%, while our GLC experiences only a 2.3\% decline.
This superior robustness stems from the LWC module, which employs a local-graph weighting mechanism to suppress the influence of noise samples and enhance representation reliability.

\vspace{+0.05cm}
\noindent
\textbf{Clustering performance on incomplete + noise setting.}
To comprehensively assess the robustness of our approach, we further conduct experiments on noise incomplete multi-view datasets, with the results summarized in Table~\ref{table:3}.
As shown in the results, most existing incomplete MVC methods suffer severe performance degradation under such dual perturbations.
In contrast, our proposed GLC consistently maintains superior performance across all datasets.
For instance, on the DHA dataset, when both the missing rate and noise rate are set to 0.5, GLC surpasses the second-best method by 7.6\% in ACC.
These results demonstrate the robustness of our framework, where the joint design of GGC and LWC enables effective learning of complementary and noise-resistant representations, leading to stable clustering under incomplete and noise conditions.

\subsection{Ablation Study.}
\textbf{Ablation study on loss components.}~From the perspective of optimization loss, our framework consists of three key components: reconstruction loss $\mathcal{L}_{\text{rec}}$, global-graph guided contrastive loss $\mathcal{L}_{\text{ggc}}$, and local-graph weighted contrastive loss $\mathcal{L}_{\text{lwc}}$.
As shown in Table~\ref{table:4}, adding $\mathcal{L}_{\text{ggc}}$ significantly improves clustering accuracy under the Incomplete setting, where the ACC on LandUse-21 increases by 8.2\%.
In the Noise setting, introducing $\mathcal{L}_{\text{lwc}}$ further enhances robustness, leading to a 4.9\% improvement in ACC on ProteinFold.
When all components are combined under the Noise-Incomplete condition, GLC achieves the best overall performance across all datasets, demonstrating the complementary and essential roles of $\mathcal{L}_{\text{ggc}}$ and $\mathcal{L}_{\text{lwc}}$ in learning robust and discriminative representations.

\begin{figure}[!t]
    \centering
    \begin{subfigure}[b]{0.235\textwidth}
        \includegraphics[width=\textwidth]{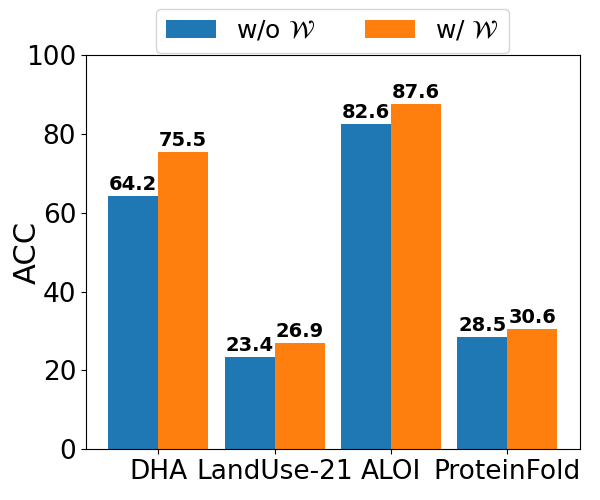}
        \caption{Incomplete setting}
    \end{subfigure}
    \hfill
    \begin{subfigure}[b]{0.235\textwidth}
        \includegraphics[width=\textwidth]{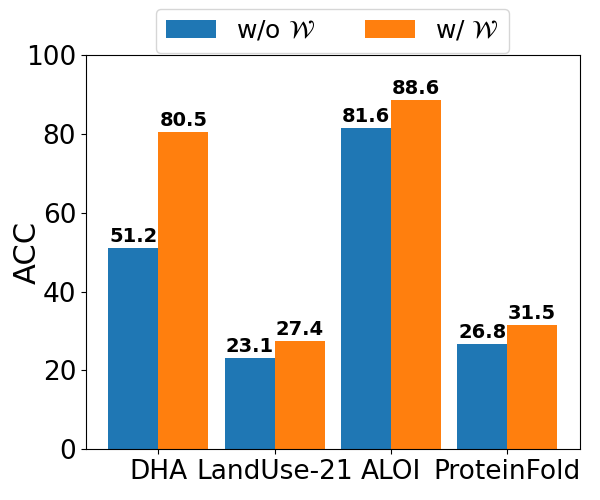}
        \caption{Noise setting}
    \end{subfigure}
    \caption{Ablation of with/without weight $\mathcal{W}$ on four datasets.}
    \label{fig:1}
\end{figure}

\vspace{+0.05cm}
\noindent
\textbf{Ablation study on weighting mechanism.}~Figure~\ref{fig:1} compares the clustering accuracy of variants with and without the proposed weighting mechanism.
In both scenarios, variant w/ $\mathcal{W}$ achieved the best performance, confirming its effectiveness in enhancing representation reliability.
For instance, in the Incomplete setting, the DHA dataset shows an ACC increase from 64.2\% to 75.5\%, while in the Noise setting, the ProteinFold dataset improves from 26.8\% to 31.5\%.
These results demonstrate that the adaptive weighting strategy effectively suppresses unreliable correspondences and strengthens the learning of stable cross-view semantics.
%-------------------------------------------------------------------------

\begin{figure}[!t]
    \centering
    \begin{subfigure}[b]{0.235\textwidth}
        \includegraphics[width=\textwidth]{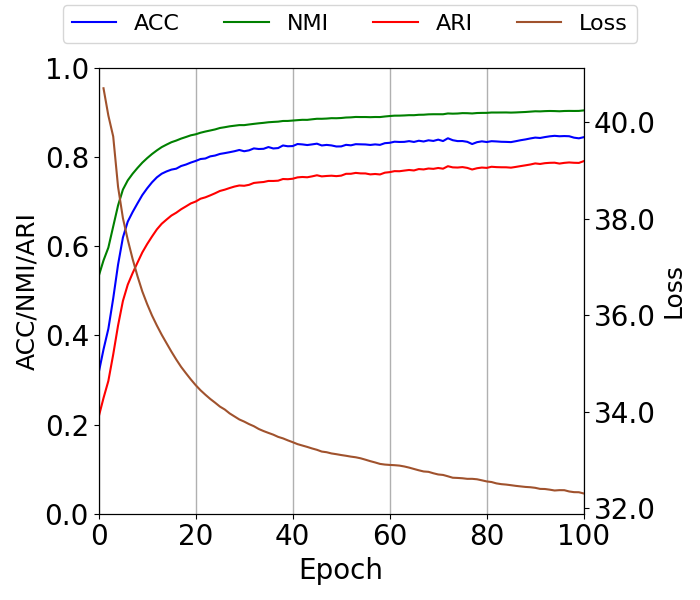}
        \caption{Incomplete setting}
    \end{subfigure}
    \hfill 
    \begin{subfigure}[b]{0.235\textwidth}
        \includegraphics[width=\textwidth]{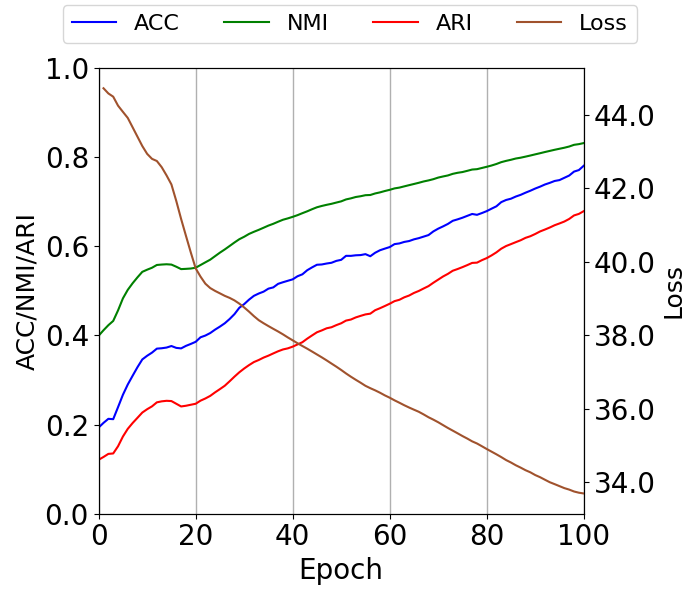}
        \caption{Noise setting}
    \end{subfigure}
    \caption{Loss $vs.$ Clustering performance on ALOI.}
    \label{fig:2}
\end{figure}

\section{Model Analysis.}
\noindent
\textbf{Training loss and performance.}
In Figure~\ref{fig:2}, we visualize the curves of loss as well as clustering performance during the model training process of GLC. 
Here, we use three clustering metrics, i.e., ACC, NMI, and Adjusted Rand Index (ARI) to measure the performance stability.
It can be observed that the loss curve exhibits a smooth and continuous downward trend, demonstrating that GLC maintains stable optimization behavior throughout the training process. 
Meanwhile, the steadily increasing curves of clustering metrics indicate that the model progressively learns more discriminative and consistent cluster structures. 
This clearly verifies the effectiveness and robustness of our optimization strategy, showing that GLC achieves reliable convergence even under incomplete and noise multi-view conditions.

\begin{figure}[!t]
    \centering
    \begin{subfigure}[b]{0.235\textwidth}
        \includegraphics[width=\textwidth]{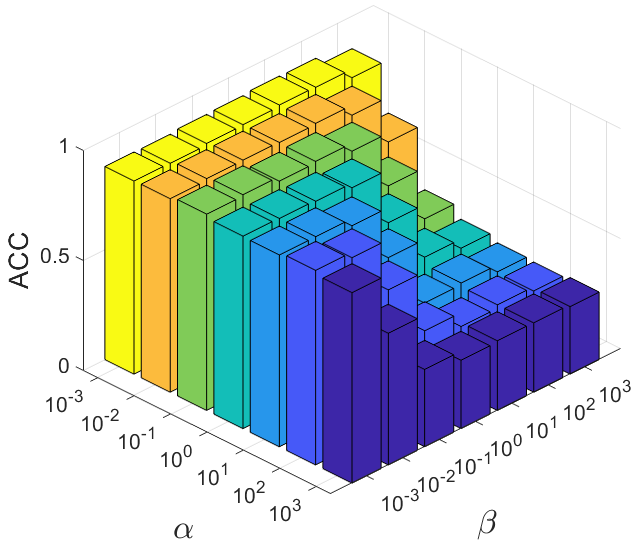}
        \caption{Incomplete setting}
    \end{subfigure}
    \hfill
    \begin{subfigure}[b]{0.235\textwidth}
        \includegraphics[width=\textwidth]{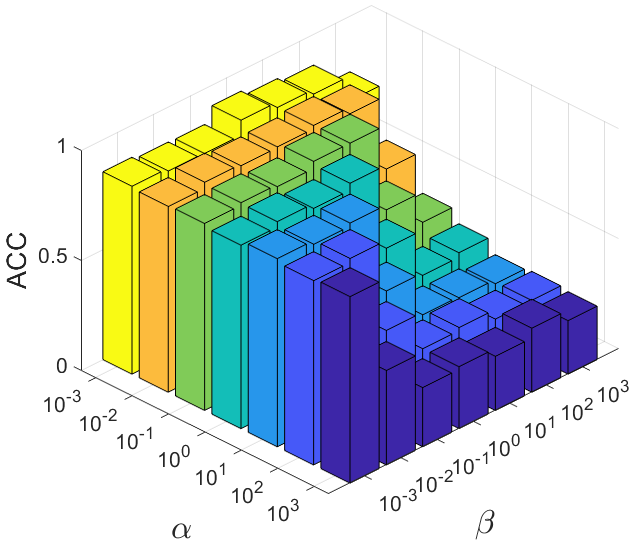}
        \caption{Noise setting}
    \end{subfigure}
    \caption{ACC \(\mathit{vs.}\) Parameters \(\{\alpha,\beta\}\) on ALOI.}
    \label{fig:3}
    \vspace{-0.2cm}
\end{figure}

\begin{figure}[!t]
    \centering
    \begin{subfigure}[b]{0.235\textwidth}
        \includegraphics[width=\textwidth]{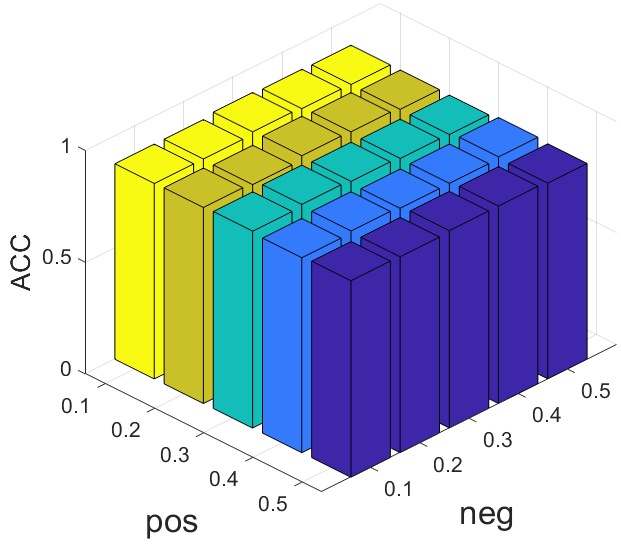}
        \caption{Incomplete setting}
    \end{subfigure}
    \hfill 
    \begin{subfigure}[b]{0.235\textwidth}
        \includegraphics[width=\textwidth]{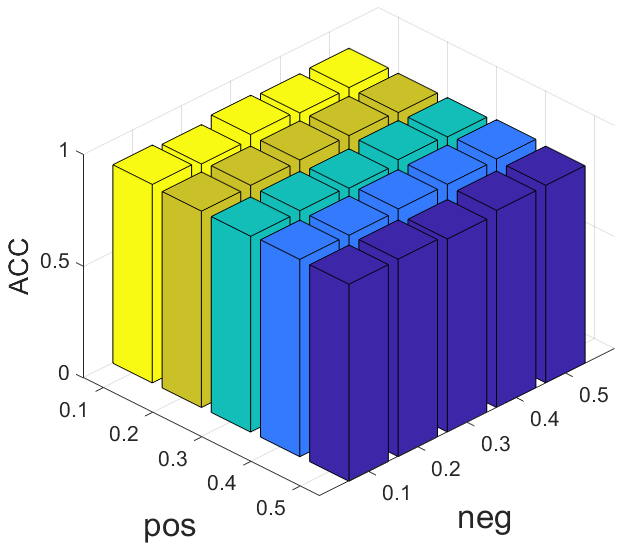}
        \caption{Noise setting}
    \end{subfigure}
    \caption{ACC \(\mathit{vs.}\) Ratios \(\{\text{pos,neg}\}\) on ALOI.}
    \label{fig:4}
\end{figure}

\vspace{+0.05cm}
\noindent
\textbf{Parameter analysis.}
In the proposed GLC framework, two hyperparameters, the loss balance weights $\alpha$ and $\beta$ in Eq.~\eqref{eq:total}, are introduced to control the relative contributions of different objectives.
Their sensitivity analysis is presented in Figure~\ref{fig:3}.
As shown, the clustering performance remains stable within a broad range of $\alpha$ and $\beta$, indicating that GLC is not sensitive to their variations.
For generality, we set $\alpha = 0.1$ and $\beta = 1.0$ for all experiments across datasets.
Furthermore, Figure~\ref{fig:4} shows the influence of the positive and negative sample ratio parameters (\textit{pos}, \textit{neg}).
The results reveal that varying these ratios has little impact on performance, confirming the robustness of our contrastive learning mechanism under different configurations.

%------------------------------------------------------------------------

\section{Conclusion}
Existing studies often handle multi-view clustering on incomplete and noise settings separately, limiting their ability to jointly exploit complementary information.
To overcome this limitation, we propose an imputation-free framework named \textbf{GLC}, which unifies robust dual contrastive learning to handle both incomplete and noise multi-view clustering tasks.
By modeling global-view semantic correlations and local-view semantic non-correlations, GLC effectively alleviates the rare-paired and mis-paired issues, leading to more reliable and noise-resilient representations.
Extensive experiments on multiple benchmark datasets demonstrate that GLC achieves robust and superior clustering performance under both incomplete and noise conditions, providing a unified solution for real-world multi-view data analysis.

{
    \small
    \bibliographystyle{ieeenat_fullname}
    \bibliography{main}

@String(AAAI = {AAAI})

@article{rappoport2018multi,
  title={Multi-omic and multi-view clustering algorithms: review and cancer benchmark},
  author={Rappoport, Nimrod and Shamir, Ron},
  journal={Nucleic acids research},
  volume={46},
  number={20},
  pages={10546--10562},
  year={2018},
  publisher={Oxford University Press}
}

@inproceedings{chao2025global,
  title={Global Graph Propagation with Hierarchical Information Transfer for Incomplete Contrastive Multi-view Clustering},
  author={Chao, Guoqing and Xu, Kaixin and Xie, Xijiong and Chen, Yongyong},
  booktitle={AAAI Conference on Artificial Intelligence},
  pages={15713--15721},
  year={2025}
}

@inproceedings{bickel2004multi,
  title={Multi-view clustering.},
  author={Bickel, Steffen and Scheffer, Tobias},
  booktitle={International Conference on Data Mining},
  pages={19--26},
  year={2004}
}

@article{wang2023self,
  title={Self-supervised image clustering from multiple incomplete views via constrastive complementary generation},
  author={Wang, Jiatai and Xu, Zhiwei and Yang, Xuewen and Guo, Dongjin and Liu, Limin},
  journal={IET Computer Vision},
  volume={17},
  number={2},
  pages={189--202},
  year={2023},
  publisher={Wiley Online Library}
}

@article{hinton2006reducing,
  title={Reducing the dimensionality of data with neural networks},
  author={Hinton, Geoffrey E and Salakhutdinov, Ruslan R},
  journal={Science},
  volume={313},
  number={5786},
  pages={504--507},
  year={2006},
  publisher={American Association for the Advancement of Science}
}

@inproceedings{xu2022deep,
  title={Deep incomplete multi-view clustering via mining cluster complementarity},
  author={Xu, Jie and Li, Chao and Ren, Yazhou and Peng, Liang and Mo, Yujie and Shi, Xiaoshuang and Zhu, Xiaofeng},
  booktitle={AAAI Conference on Artificial Intelligence},
  pages={8761--8769},
  year={2022}
}

@article{xu2023self,
  title={Self-weighted contrastive learning among multiple views for mitigating representation degeneration},
  author={Xu, Jie and Chen, Shuo and Ren, Yazhou and Shi, Xiaoshuang and Shen, Hengtao and Niu, Gang and Zhu, Xiaofeng},
  journal={Advances in Neural Information Processing Systems},
  volume={36},
  pages={1119--1131},
  year={2023}
}

@inproceedings{vazquez2020multigraph,
  title={Multigraph spectral clustering for joint content delivery and scheduling in beam-free satellite communications},
  author={V{\'a}zquez, Miguel Angel and P{\'e}rez-Neira, Ana I},
  booktitle={International Conference on Acoustics, Speech, and Signal Processing},
  pages={8802--8806},
  year={2020}
}

@inproceedings{yu2018web,
  title={Web items recommendation based on multi-view clustering},
  author={Yu, Hong and Zhang, Tiantian and Chen, Jiaxin and Guo, Chen and Lian, Yahong},
  booktitle={International Conference on Computers, Software, and Applications},
  pages={420--425},
  year={2018},
}

@inproceedings{lin2021completer,
  title={Completer: Incomplete multi-view clustering via contrastive prediction},
  author={Lin, Yijie and Gou, Yuanbiao and Liu, Zitao and Li, Boyun and Lv, Jiancheng and Peng, Xi},
  booktitle={Computer Vision and Pattern Recognition Conference},
  pages={11174--11183},
  year={2021}
}

@inproceedings{jin2023deep,
  title={Deep incomplete multi-view clustering with cross-view partial sample and prototype alignment},
  author={Jin, Jiaqi and Wang, Siwei and Dong, Zhibin and Liu, Xinwang and Zhu, En},
  booktitle={IEEE/CVF Conference on Computer Vision and Pattern Recognition},
  pages={11600--11609},
  year={2023}
}

@inproceedings{xu2024deep,
  title={Deep variational incomplete multi-view clustering: Exploring shared clustering structures},
  author={Xu, Gehui and Wen, Jie and Liu, Chengliang and Hu, Bing and Liu, Yicheng and Fei, Lunke and Wang, Wei},
  booktitle={AAAI Conference on Artificial Intelligence},
  pages={16147--16155},
  year={2024}
}

@article{liu2018late,
  title={Late fusion incomplete multi-view clustering},
  author={Liu, Xinwang and Zhu, Xinzhong and Li, Miaomiao and Wang, Lei and Tang, Chang and Yin, Jianping and Shen, Dinggang and Wang, Huaimin and Gao, Wen},
  journal={IEEE Transactions on Pattern Analysis and Machine Intelligence},
  volume={41},
  number={10},
  pages={2410--2423},
  year={2018},
  publisher={IEEE}
}

@inproceedings{wang2020robust,
  title={Robust self-weighted multi-view projection clustering},
  author={Wang, Beilei and Xiao, Yun and Li, Zhihui and Wang, Xuanhong and Chen, Xiaojiang and Fang, Dingyi},
  booktitle={AAAI conference on artificial intelligence},
  volume={34},
  number={04},
  pages={6110--6117},
  year={2020}
}

@article{wang2025deep,
  title={Deep incomplete multi-view clustering via multi-level imputation and contrastive alignment},
  author={Wang, Ziyu and Du, Yiming and Wang, Yao and Ning, Rui and Li, Lusi},
  journal={Neural Networks},
  volume={181},
  pages={106851},
  year={2025},
  publisher={Elsevier}
}

@inproceedings{wu2025imputation,
  title={Imputation-free incomplete multi-view clustering via knowledge distillation},
  author={Wu, Benyu and Du, Wei and Wang, Jun and Yu, Guoxian},
  booktitle={IEEE/CVF Conference on Computer Vision and Pattern Recognition},
  pages={5071--5081},
  year={2025}
}

@inproceedings{chaudhuri2009multi,
  title={Multi-view clustering via canonical correlation analysis},
  author={Chaudhuri, Kamalika and Kakade, Sham M and Livescu, Karen and Sridharan, Karthik},
  booktitle={International Conference on Machine Learning},
  pages={129--136},
  year={2009}
}

@article{li2023scbridge,
  title={scBridge embraces cell heterogeneity in single-cell RNA-seq and ATAC-seq data integration},
  author={Li, Yunfan and Zhang, Dan and Yang, Mouxing and Peng, Dezhong and Yu, Jun and Liu, Yu and Lv, Jiancheng and Chen, Lu and Peng, Xi},
  journal={Nature Communications},
  volume={14},
  number={1},
  pages={6045},
  year={2023},
  publisher={Nature Publishing Group UK London}
}

@inproceedings{yuan2024robust,
  title={Robust prototype completion for incomplete multi-view clustering},
  author={Yuan, Honglin and Lai, Shiyun and Li, Xingfeng and Dai, Jian and Sun, Yuan and Ren, Zhenwen},
  booktitle={ACM International Conference on Multimedia},
  pages={10402--10411},
  year={2024}
}

@article{li2020bipartite,
  title={Bipartite graph based multi-view clustering},
  author={Li, Lusi and He, Haibo},
  journal={IEEE Transactions on Knowledge and Data Engineering},
  volume={34},
  number={7},
  pages={3111--3125},
  year={2020},
  publisher={IEEE}
}

@inproceedings{lou2024self,
  title={Self-supervised weighted information bottleneck for multi-view clustering},
  author={Lou, Zhengzheng and Zhang, Chaoyang and Xue, Hang and Ye, Yangdong and Zhou, Qinglei and Hu, Shizhe},
  booktitle={International Joint Conference on Artificial Intelligence},
  pages={4643--4650},
  year={2024}
}

@article{paszke2019pytorch,
  title={Pytorch: An imperative style, high-performance deep learning library},
  author={Paszke, Adam and Gross, Sam and Massa, Francisco and Lerer, Adam and Bradbury, James and Chanan, Gregory and Killeen, Trevor and Lin, Zeming and Gimelshein, Natalia and Antiga, Luca and others},
  journal={Advances in Neural Information Processing Systems},
  volume={32},
  year={2019}
}

@inproceedings{yang2023dealmvc,
  title={Dealmvc: Dual contrastive calibration for multi-view clustering},
  author={Yang, Xihong and Jiaqi, Jin and Wang, Siwei and Liang, Ke and Liu, Yue and Wen, Yi and Liu, Suyuan and Zhou, Sihang and Liu, Xinwang and Zhu, En},
  booktitle={ACM international conference on multimedia},
  pages={337--346},
  year={2023}
}

@inproceedings{trosten2021reconsidering,
  title={Reconsidering representation alignment for multi-view clustering},
  author={Trosten, Daniel J and Lokse, Sigurd and Jenssen, Robert and Kampffmeyer, Michael},
  booktitle={IEEE/CVF Conference on Computer Vision and Pattern Recognition},
  pages={1255--1265},
  year={2021}
}

@inproceedings{liu2023adaptive,
  title={Adaptive weighted multi-view clustering},
  author={Liu, Shuo Shuo and Lin, Lin},
  booktitle={Conference on Health, Inference, and Learning},
  pages={19--36},
  year={2023},
  organization={PMLR}
}

@inproceedings{xu2022multi,
  title={Multi-level feature learning for contrastive multi-view clustering},
  author={Xu, Jie and Tang, Huayi and Ren, Yazhou and Peng, Liang and Zhu, Xiaofeng and He, Lifang},
  booktitle={IEEE/CVF Conference on Computer Vision and Pattern Recognition},
  pages={16051--16060},
  year={2022}
}

@article{cui2024dual,
  title={Dual contrast-driven deep multi-view clustering},
  author={Cui, Jinrong and Li, Yuting and Huang, Han and Wen, Jie},
  journal={IEEE Transactions on Image Processing},
  year={2024},
  publisher={IEEE}
}

@article{wang2021generative,
  title={Generative partial multi-view clustering with adaptive fusion and cycle consistency},
  author={Wang, Qianqian and Ding, Zhengming and Tao, Zhiqiang and Gao, Quanxue and Fu, Yun},
  journal={IEEE Transactions on Image Processing},
  volume={30},
  pages={1771--1783},
  year={2021},
  publisher={IEEE}
}

@inproceedings{chen2023deep,
  title={Deep multiview clustering by contrasting cluster assignments},
  author={Chen, Jie and Mao, Hua and Woo, Wai Lok and Peng, Xi},
  booktitle={IEEE/CVF Conference on Computer Vision and Pattern Recognition},
  pages={16752--16761},
  year={2023}
}

@inproceedings{lin2012human,
  title={Human action recognition and retrieval using sole depth information},
  author={Lin, Yan-Ching and Hu, Min-Chun and Cheng, Wen-Huang and Hsieh, Yung-Huan and Chen, Hong-Ming},
  booktitle={ACM International Conference on Multimedia},
  pages={1053--1056},
  year={2012}
}

@inproceedings{tang2022deep,
  title={Deep safe incomplete multi-view clustering: Theorem and algorithm},
  author={Tang, Huayi and Liu, Yong},
  booktitle={International Conference on Machine Learning},
  pages={21090--21110},
  year={2022},
  organization={PMLR}
}

@inproceedings{wen2020dimc,
  title={Dimc-net: Deep incomplete multi-view clustering network},
  author={Wen, Jie and Zhang, Zheng and Zhang, Zhao and Wu, Zhihao and Fei, Lunke and Xu, Yong and Zhang, Bob},
  booktitle={ACM international conference on multimedia},
  pages={3753--3761},
  year={2020}
}

@inproceedings{pu2024adaptive,
  title={Adaptive feature imputation with latent graph for deep incomplete multi-view clustering},
  author={Pu, Jingyu and Cui, Chenhang and Chen, Xinyue and Ren, Yazhou and Pu, Xiaorong and Hao, Zhifeng and Yu, Philip S and He, Lifang},
  booktitle={AAAI conference on artificial intelligence},
  volume={38},
  number={13},
  pages={14633--14641},
  year={2024}
}

@inproceedings{yang2010bag,
  title={Bag-of-visual-words and spatial extensions for land-use classification},
  author={Yang, Yi and Newsam, Shawn},
  booktitle={SIGSPATIAL International Conference on Advances in Geographic Information Systems},
  pages={270--279},
  year={2010}
}

@article{damoulas2008probabilistic,
  title={Probabilistic multi-class multi-kernel learning: on protein fold recognition and remote homology detection},
  author={Damoulas, Theodoros and Girolami, Mark A},
  journal={Bioinformatics},
  volume={24},
  number={10},
  pages={1264--1270},
  year={2008},
  publisher={Oxford University Press}
}

@article{lin2022dual,
  title={Dual contrastive prediction for incomplete multi-view representation learning},
  author={Lin, Yijie and Gou, Yuanbiao and Liu, Xiaotian and Bai, Jinfeng and Lv, Jiancheng and Peng, Xi},
  journal={IEEE Transactions on Pattern Analysis and Machine Intelligence},
  volume={45},
  number={4},
  pages={4447--4461},
  year={2022},
  publisher={IEEE}
}

@article{kingma2014adam,
  title={Adam: A method for stochastic optimization},
  author={Kingma, Diederik P},
  journal={arXiv preprint arXiv:1412.6980},
  year={2014}
}

@inproceedings{liu2024sample,
  title={Sample-level cross-view similarity learning for incomplete multi-view clustering},
  author={Liu, Suyuan and Zhang, Junpu and Wen, Yi and Yang, Xihong and Wang, Siwei and Zhang, Yi and Zhu, En and Tang, Chang and Zhao, Long and Liu, Xinwang},
  booktitle={AAAI Conference on Artificial Intelligence},
  pages={14017--14025},
  year={2024}
}

@inproceedings{mcqueen1967some,
  title={Some methods of classification and analysis of multivariate observations},
  author={McQueen, James B},
  booktitle={Proc. of 5th Berkeley Symposium on Math. Stat. and Prob.},
  pages={281--297},
  year={1967}
}

@article{tian2020makes,
  title={What makes for good views for contrastive learning?},
  author={Tian, Yonglong and Sun, Chen and Poole, Ben and Krishnan, Dilip and Schmid, Cordelia and Isola, Phillip},
  journal={Advances in Neural Information Processing Systems},
  volume={33},
  pages={6827--6839},
  year={2020}
}

@article{xi2025contrastive,
  title={Contrastive and Dual Adversarial Representation Learning for Multi-view Clustering},
  author={Xi, Yanwanyu and Tang, Chang and Huang, Jun-Jie and Hu, Xingchen and Liu, Yuanyuan and Liu, Xinwang},
  journal={IEEE Transactions on Knowledge and Data Engineering},
  year={2025},
  publisher={IEEE}
}

@inproceedings{dai2025imputation,
  title={Imputation-free and Alignment-free: Incomplete Multi-view Clustering Driven by Consensus Semantic Learning},
  author={Dai, Yuzhuo and Jin, Jiaqi and Dong, Zhibin and Wang, Siwei and Liu, Xinwang and Zhu, En and Yang, Xihong and Gan, Xinbiao and Feng, Yu},
  booktitle={Computer Vision and Pattern Recognition Conference},
  pages={5071--5081},
  year={2025}
}

@inproceedings{xu2024investigating,
  title={Investigating and mitigating the side effects of noisy views for self-supervised clustering algorithms in practical multi-view scenarios},
  author={Xu, Jie and Ren, Yazhou and Wang, Xiaolong and Feng, Lei and Zhang, Zheng and Niu, Gang and Zhu, Xiaofeng},
  booktitle={Proceedings of the IEEE/CVF conference on computer vision and pattern recognition},
  pages={22957--22966},
  year={2024}
}

@article{du2021deep,
  title={Deep multiple auto-encoder-based multi-view clustering},
  author={Du, Guowang and Zhou, Lihua and Yang, Yudi and L{\"u}, Kevin and Wang, Lizhen},
  journal={Data Science and Engineering},
  volume={6},
  number={3},
  pages={323--338},
  year={2021},
  publisher={Springer}
}

@inproceedings{zhang2025incomplete,
  title={Incomplete Multi-view Clustering via Diffusion Contrastive Generation},
  author={Zhang, Yuanyang and Lin, Yijie and Yan, Weiqing and Yao, Li and Wan, Xinhang and Li, Guangyuan and Zhang, Chao and Ke, Guanzhou and Xu, Jie},
  booktitle={AAAI Conference on Artificial Intelligence},
  pages={22650--22658},
  year={2025}
}
}

\end{document}